# Clinical semantics for lung cancer prediction

Luis H. John[1], Jan A. Kors[1], Jenna M. Reps[1,2], Peter R. Rijnbeek[1], Egill A. Fridgeirsson[1]

[1]Department of Medical Informatics, Erasmus University Medical Center, Rotterdam, Netherlands

[2]Johnson & Johnson, Titusville, NJ, United States

**Abstract**

Background: Existing clinical prediction models often represent patient data using features that ignore the semantic relationships between clinical concepts. This study integrates domain-specific semantic information by mapping the SNOMED medical term hierarchy into a low-dimensional hyperbolic space using Poincaré embeddings, with the aim of improving lung cancer onset prediction.

Methods: Using a retrospective cohort from the Optum EHR dataset, we derived a clinical knowledge graph from the SNOMED taxonomy and generated Poincaré embeddings via Riemannian stochastic gradient descent. These embeddings were then incorporated into two deep learning architectures, a ResNet and a Transformer model. Models were evaluated for discrimination (area under the receiver operating characteristic curve) and calibration (average absolute difference between observed and predicted probabilities) performance.

Results: Incorporating pre-trained Poincaré embeddings resulted in modest and consistent improvements in discrimination performance compared to baseline models using randomly initialized Euclidean embeddings. ResNet models, particularly those using a 10-dimensional Poincaré embedding, showed enhanced calibration, whereas Transformer models maintained stable calibration across configurations.

Discussion: Embedding clinical knowledge graphs into hyperbolic space and integrating these representations into deep learning models can improve lung cancer onset prediction by preserving the hierarchical structure of clinical terminologies used for prediction. This approach demonstrates a feasible method for combining data-driven feature extraction with established clinical knowledge.

## 1 Background

Prediction of disease onset can guide early intervention and improve patient outcomes.(1) Clinical predictive models typically rely on standard feature encodings such as one-hot encoded categorical variables or binary indicators of clinical events, which treat each clinical concept as an independent variable.(2, 3) This approach neglects the inherent semantic relationships and hierarchical structures that exist within clinical terminologies.(4) In this work we aim to integrate domain-specific semantic information from clinical taxonomies into clinical prediction models for lung cancer, which is the leading cause of cancer mortality in the United States.(5). We use Poincaré embeddings to map clinical concepts from a medical terms hierarchy into a hyperbolic low-dimensional latent space, thereby preserving the tree-like structure of these terminologies and capturing inter-concept relationships that standard encodings fail to represent.(6) We integrate the pre-trained embeddings into both ResNet and Transformer models, which are evaluated on a large observational healthcare dataset.(7) This approach bridges the gap between data-driven feature selection and expert-curated clinical knowledge, to potentially yield more robust and more interpretable predictions on large observational health data.

## 2  Methods

Our methodological framework includes two core steps: (1) generating Poincaré embeddings from a clinical knowledge graph, and (2) integrating these embeddings into deep learning architectures for patient-level prediction.

### 2.1  Data and Prediction Problem

This retrospective study uses Optum® de-identified Electronic Health Record dataset (Optum EHR), a structured observational healthcare dataset from Optum's longitudinal electronic health record (EHR) repository which is derived from dozens of healthcare provider organizations in the United States. The data is certified as de-identified by an independent statistical expert following HIPAA statistical de-identification rules and managed according to Optum customer data use agreements. Clinical, claims and other medical administrative data is obtained from both inpatient and ambulatory EHRs, practice management systems and numerous other internal systems from where information is processed, normalized, and standardized. As a final step, this database is mapped to the Observational Medical Outcomes Partnership (OMOP) Common Data Model (CDM) to facilitate the development of patient-level prediction models best practices established by the Observational Health Data Science and Informatics (OHDSI) initiative.(8) Optum EHR contains 111.4 million patient records in a time period from 01/2007 – 12/2022.

A patient-level prediction model quantifies (i.e. predicts) a person's risk of developing a health outcome during a specified time-at-risk period following an index date, using information collected in an observation window prior to index (Figure 1). The health outcome of interest is lung cancer, which is the leading cause of cancer mortality in the United States, with a 5-year survival rate of only 22% due to often late-stage diagnosis.(5) Despite the proven benefits of screening, uptake is low, and many patients diagnosed with lung cancer do not meet current screening criteria.(9-11) Early detection of lung cancer can significantly improve treatment outcomes and survival rates. In this study, we predict lung cancer in a target cohort of persons aged 45 – 65 during a time-at-risk of 1,095 days as defined in a published clinical article.(12)

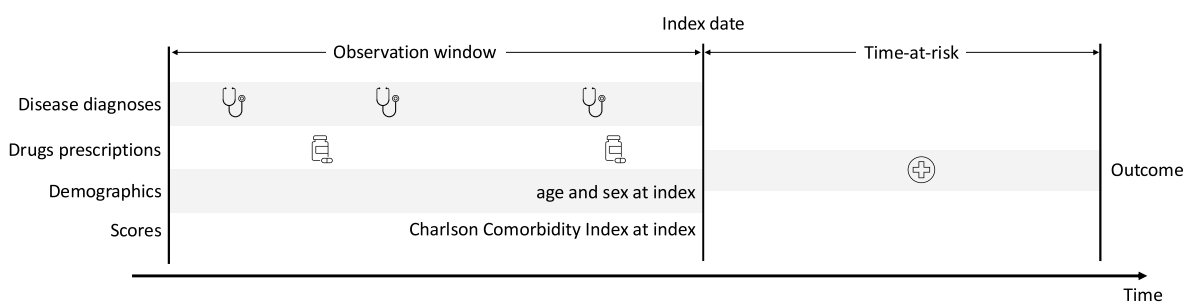

Figure 1. Lung cancer onset prediction time windows for a person in the study population.(13)

For lung cancer prediction, a visit record that marks an interaction with a healthcare provider, serves as the index date, allowing for the practical application of the model. To use recent data, but at the same time eliminate pandemic-related confounding effects on healthcare systems and patient behavior, we chose to utilize pre-COVID data from before the pandemic (before 1 January 2020). Given the time-at-risk period this means that the index date for lung cancer falls into the period of 1 Jan 2016 – 31 Dec 2016.

Participants require 365 days of continuous observation time before the index date, in which candidate predictors are assessed (Figure 1). This relatively short period is consistent with other models in literature and, as opposed to all-time lookback, was also found to have only small impact on discrimination and calibration as all-time lookback can vary strongly across patients.(14) As candidate predictors, we use a patient's age, sex, and Charlson Comorbidity Index at the index date. During the observation window we use dichotomized diagnoses and drug prescriptions. Even though this information is recorded at multiple time points, for analysis purposes it is flattened into a tabular format. Additionally, following empirical recommendations on handling patients lost to follow-up, the study allows participants to exit the cohorts at any time during the time-at-risk period, provided they have at least one day of time-at-risk after the index date.(15)

## 2.2 Poincaré Embeddings of Clinical Knowledge Graphs

We use a knowledge graph based on a clinical taxonomy, the clinical terms ("clinical findings") hierarchy from the Systematic Nomenclature of Medicine (SNOMED), maintained by the International Health Terminology Standards Development Organisation (IHTSDO).(16) For our purposes, we use the OHDSI vocabulary (version 20240830) of SNOMED as implemented in the OMOP CDM. The concept hierarchy is provided as an edge list, in which nodes represent clinical concepts that subsume other concepts.

We identify all clinical findings recorded during the lung cancer observation window and treat them as candidate predictors. Because the SNOMED taxonomy covers only clinical findings, we extract the ancestral subtree from the full SNOMED hierarchy that contains each finding plus every intermediate finding on the path from a root to that finding. Thus, any SNOMED concept not observed in our data is excluded. We preserve multiple-parent relationships so that each finding may have more than one ancestor.

All candidate predictors drawn from this SNOMED subtree are embedded using Poincaré embeddings to capture the hierarchy. Any other covariates recorded in the observation window such as drug prescriptions that fall outside the SNOMED taxonomy are processed separately.

To represent this hierarchical data, we adopt hyperbolic embeddings following Nickel and Kiela's approach.(6) Hierarchical data naturally forms a tree structure whose number of nodes grows exponentially with depth. In Euclidean space, capturing such exponential growth typically requires high-dimensional representation, leading to potential overfitting and increased computational cost.(6) In contrast, in the Poincaré ball model, space expands toward the boundary. A small increase in radius near the center adds only modest volume, but the same increase near the edge adds a large volume. As a result, ball volume grows roughly exponentially with radius, mirroring branching of a tree and allowing deep hierarchies to fit in low-dimensional Poincaré models.

We embed the data in the Poincaré ball defined as

$$\mathcal{B}^n = \{x \in \mathbb{R}^n : ||x|| < 1\}$$

And equipped with the Riemannian metric

$$g_x = \left(\frac{2}{1 - ||x||^2}\right)^2 g^E$$

where $g^E$ denotes the Euclidean metric. In this space, the distance between two points $u, v \in \mathcal{B}^n$ is computed as

$$d(u,v) = \operatorname{arcosh}\left(1 + 2\frac{||u-v||^2}{(1-||u||^2)(1-||v||^2)}\right)$$

Because hyperbolic space reflects exponential growth with respect to distance, it naturally accommodates the exponential branching of tree-like data in a low-dimensional manifold. This formulation preserves constant geodesic lengths between connected nodes in the original hierarchy and faithfully captures the intrinsic structure of this clinical taxonomy.

## 2.3 Training and Optimization

We learn the Poincaré embeddings using Riemannian stochastic gradient descent (RSGD) adapted to the hyperbolic manifold. At each epoch, embedding coordinates are updated using gradients computed with respect to the Riemannian metric, thereby ensuring that the embedding remains in the open unit ball.(6)

The supervisory signal during training comes from the presence or absence of edges in the SNOMED clinical terms hierarchy. To promote convergence, all nodes are initialized near the origin of the Poincaré ball.(6)

As described by Nickel and Kiela, let $\mathcal{D} = \{(u,v)\}$ be the set of parent-child relations between concept pairs which will also be referred to as positive edges. (6) We learn embeddings of all symbols in $\mathcal{D}$ such that related objects are close in the embedding space through minimizing the loss function

$$\mathcal{L}(\Theta) = \sum_{(u,v)\in\mathcal{D}} \log \frac{e^{-d(u,v)}}{\sum_{v'\in\mathcal{N}(u)} e^{-d(u,v')}},$$

where $\mathcal{N}(u) = \{v | (u,v) \notin \mathcal{D}\} \cup \{u\}$ is the set of negative examples for $u$ (including $u$), referred to as negative edges. The nodes that do not have an edge connecting them will be referred to as negative neighbors. (6) Similarly, if a subsume relationship between two concepts in the SNOMED taxonomy exists, the edge between those two nodes will be referred to as positive edge.

We explore a hyperparameter space comprising the following factors:

- Embedding dimensions: 3, 10, 30, 100 dimensions for the Poincaré embedding.
- Burn-in epochs: 10 or 100 epochs during which the learning rate is reduced by a factor of 10 at the outset.
- Negative neighbours: 10, 50, or 100, representing the number of negative edges assessed alongside each positive edge.
- Directedness: A Boolean setting indicating whether the edge list is treated as directed (i.e., if an edge A→B exists, the reverse edge B→A is considered a negative example) or undirected.

The primary objective of this parameter exploration is to optimize node proximity in the embedding space. The learned embeddings are then optimized by minimizing the loss function and evaluated using the mean rank metric, which is the average position of the true item in a model's ranked list of candidates where lower values mean better ranking.

The embeddings are computed under an undirected assumption meaning that the presence of an edge in one direction suffices for evaluation even if the reverse edge is absent.

## 2.4 Integration with Deep Learning Architectures

Once the clinical concepts are embedded in the Poincaré model, the resulting embedding is incorporated into patient-level prediction models. For each patient, we map every recorded SNOMED code to its Poincaré vector. In the ResNet, which cannot handle sequence data, we average the several patient's vectors into a single vector. In the Transformer, we treat each Poincaré vector as a token in a sequence. We pad each patient's concept code list to the cohort's maximum length, compute the Poincaré embedding for each code, and pass the resulting sequence to the downstream model layers.

We integrate the Poincaré embeddings into two distinct deep learning models adapted for tabular data, one based on ResNet and another on the Transformer architecture.(7) Originally, these two models have randomly initilaized Euclidean embedding of 256 and 192 dimensions, respectively, which are trained alongside the model. Only concepts part of the SNOMED taxonomy can be represented in the Poincaré embedding. The remaining concept, which mainly includes drug prescriptions, but also a few concepts that are part of the standard condition covariate set of the OMOP CDM FeatureExtraction R package but not part of the SNOMED taxonomy, are instead modelled using the Euclidean embeddings.

Hyperbolic embeddings exist on a curved manifold, so we cannot feed them directly into standard neural network layers that assume Euclidean geometry. To account for this, we use a logarithmic map to project each Poincaré vector onto the Euclidean tangent plane which preserves the embedding's hierarchical relationships. We also freeze the original Poincaré embedding layer and stack an additional trainable embedding layer that learns additive offsets for task-specific fine-tuning.

Models are evaluated on a test set for discrimination using the area under the receiver operating characteristic curve, and for calibration using the average absolute difference between observed and predicted probabilities. A complete study overview is presented in Figure 2.

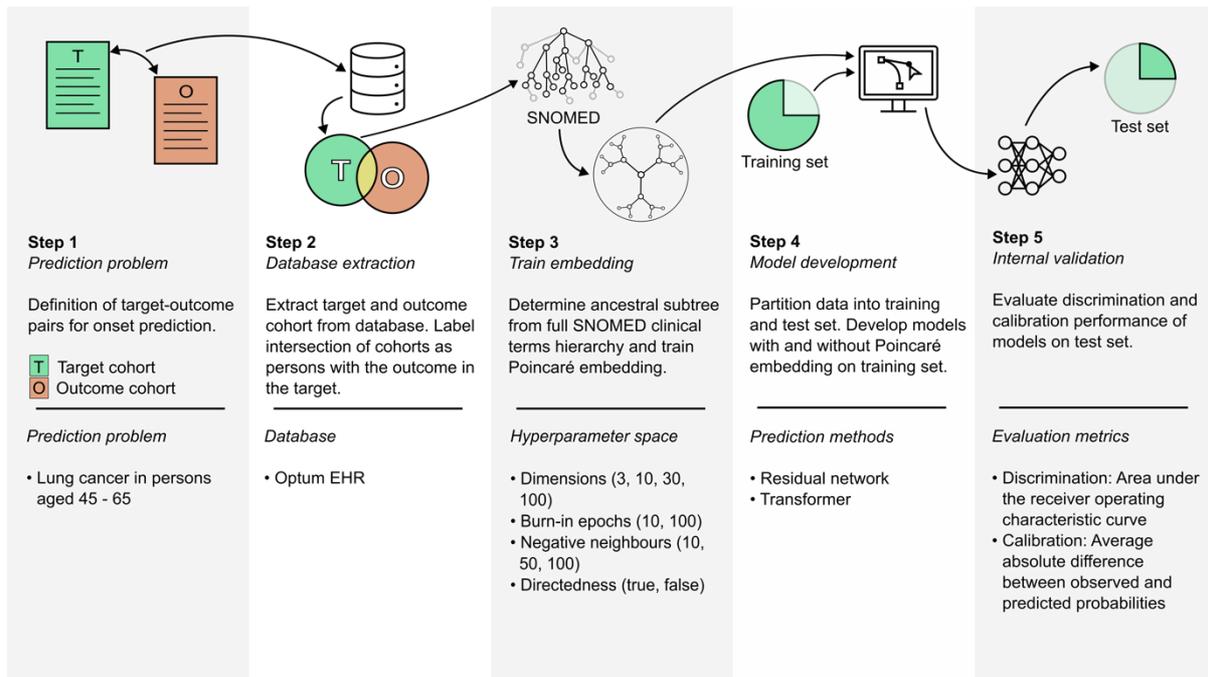

**Figure 2. Study overview.**

## 3 Results

### 3.1 Poincaré Embedding

We evaluate the mean rank (where lower is better) of the Poincaré embeddings across various hyperparameter combinations. Figure 3 presents the performance of embeddings generated using a directed SNOMED taxonomy graph. The SNOMED taxonomy uses a directed subsume relationship between a parent concept and its child. However, for the purpose of embedding concepts we strive for closeness of related concepts and directedness is not of importance. Therefore, we also assess an undirected SNOMED taxonomy graph for the embedding (refer to Appendix A). We find that the directed graph yields consistent results with the undirected graph and adopt the configuration of directedness as it represents the original form of the SNOMED taxonomy. The results indicate that the mean rank decreases primarily with increasing embedding dimensions. The number of burn-in epoch did not show a consistent effect on mean rank, whereas relying on a larger number of negative neighbors showed marginal improvements of the mean rank metric for the higher-dimensional embeddings. Based on these findings, we selected the four optimal embeddings corresponding to dimensions of 3, 10, 30, and 100, each built with 10 burn-in epochs and 100 negative neighbours. The lung cancer dataset includes a total of 26906 covariates, which among others include demographics, condition diagnoses and drug prescriptions. The full SNOMED taxonomy contains 545821 nodes and 760951 edges. The final ancestral subtree to be embedded which contains the concepts of the dataset as well as intermediate concepts from the taxonomy includes 38351 nodes. These embeddings were subsequently used for the analysis in patient-level prediction models.

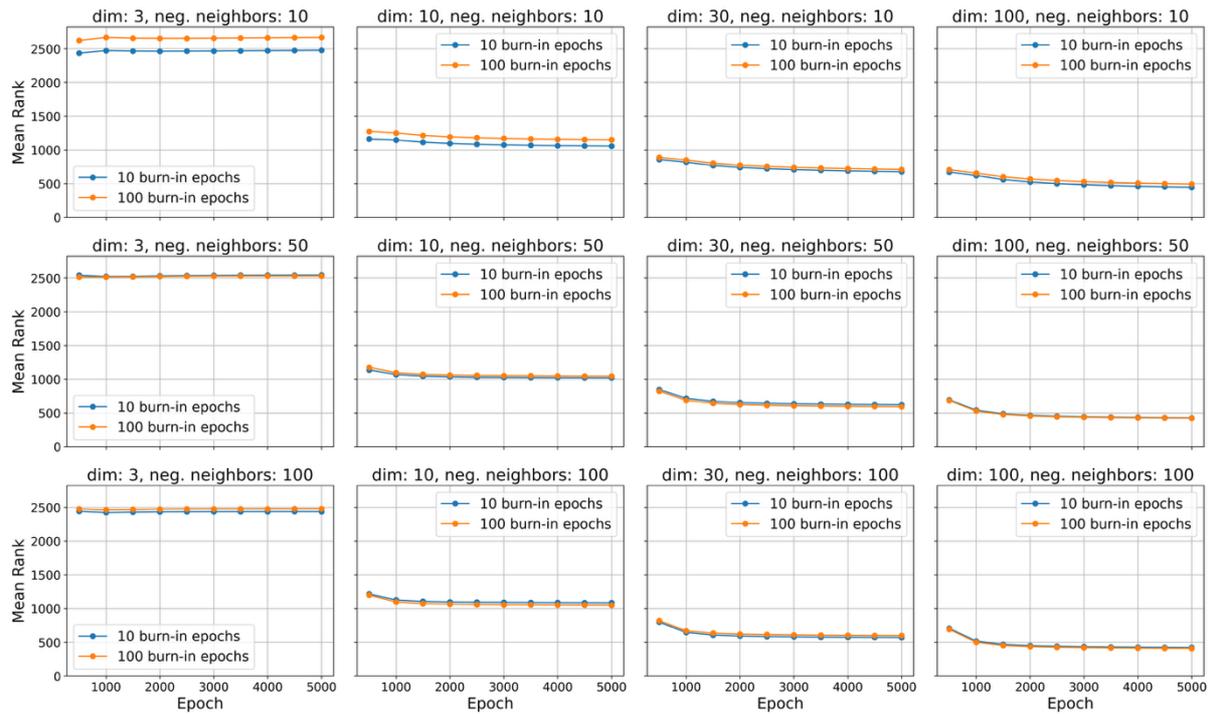

**Figure 3. Poincaré embeddings trained on directed SNOMED graph for the hyperparameter space explored.**

## 3.2 Model performance

We developed ResNet and Transformer models for each of the four selected pre-trained Poincaré embeddings and compared them to baseline models that use randomly initialized, trainable Euclidean embedding layers (with dimensions of 256 for ResNet and 192 for Transformer).

For lung cancer prediction, both ResNet and Transformer with pre-trained Poincaré embeddings achieved a modest and consistent improvement up to 0.72 and 0.70 AUROC, respectively, in discrimination performance compared to their baseline counterparts which achieved 0.70 and 0.67 AUROC, respectively (Figure 4).

For calibration, while the ResNet models benefit from pre-trained Poincaré embeddings it appears to be by a minor margin. Similarly, the calibration of the Transformer models remained largely stable overall considering overall excellent calibration, though the 10-dimensional embedding was associated with a noticeable yet minor decline.

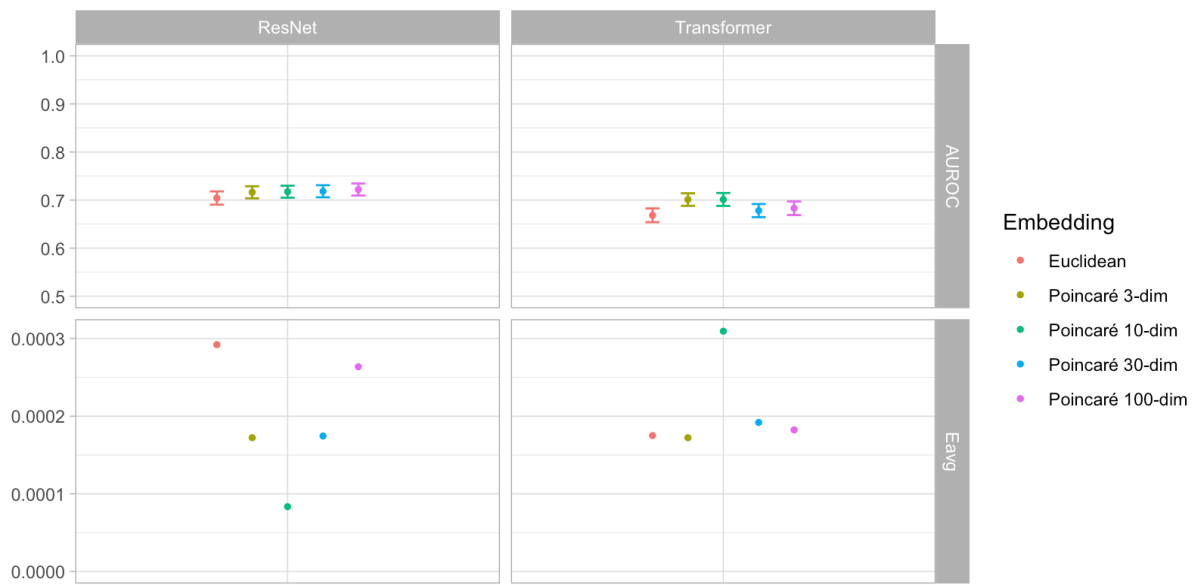

Figure 4. Discrimination (AUROC with 95% confidence intervals) and calibration $E_{avg}$ performance for models with baseline Euclidean and Poincaré embeddings.

## 4  Discussion

This study integrates domain-specific semantic information using hyperbolic Poincaré embeddings into deep learning models for lung cancer onset prediction. By converting a hierarchical representation of clinical concepts from SNOMED into a low-dimensional manifold, our approach preserves the tree-like relationships inherent in clinical terminologies, which traditional one-hot or binary encodings fail to capture.(6, 16)

Incorporating these pre-trained embeddings into both ResNet and Transformer models resulted in a modest but consistent improvement in discrimination performance. The minor improvements in calibration may not have practical significance and possibly are the effect of random noise and variation. Regardless, calibration was found to be excellent across all models trained.

An advantage of our approach is its ability to combine data-driven feature extraction with established clinical knowledge without relying on complex feature engineering. We reckon that the approach can be scaled to more prediction problems by including Poincaré concept embeddings of additional and commonly seen concepts. Concepts not in the Poincaré embedding will default to use the Euclidean embedding as shown in this work. This means that Poincaré embeddings can be made available and effectively reused.

However, while incorporating Poincaré embeddings preserves the relationships from SNOMED taxonomies, it doesn't automatically make the overall deep learning model interpretable. In other words, although the distance between embedded concepts can be examined to see how clinical concepts relate hierarchically, the predictive mechanisms within deep learning models like ResNet or Transformer remain largely opaque.

Moreover, our analysis is limited to a single large observational database and a single prediction task, and future work is needed to assess generalizability across different populations and external databases. We hypothesize that our pre-trained embeddings may affect transportability of the models positively as they could address some degree of database heterogeneity. More complex knowledge graphs such as ontologies that contain etiological information may represent an interesting area of future research in which clinical knowledge could inform the prediction model about cause or origin of disease. Also more appropriate models may exist in the form of a graph neural network to directly utilize knowledge graph information for prediction.(17)

We find that embedding clinical knowledge graphs into hyperbolic space and incorporating these representations into deep learning models offers potential advantages. Modest improvement was found for a single use case and further evaluation on other databases and outcomes is needed.

# 6 Appendix A

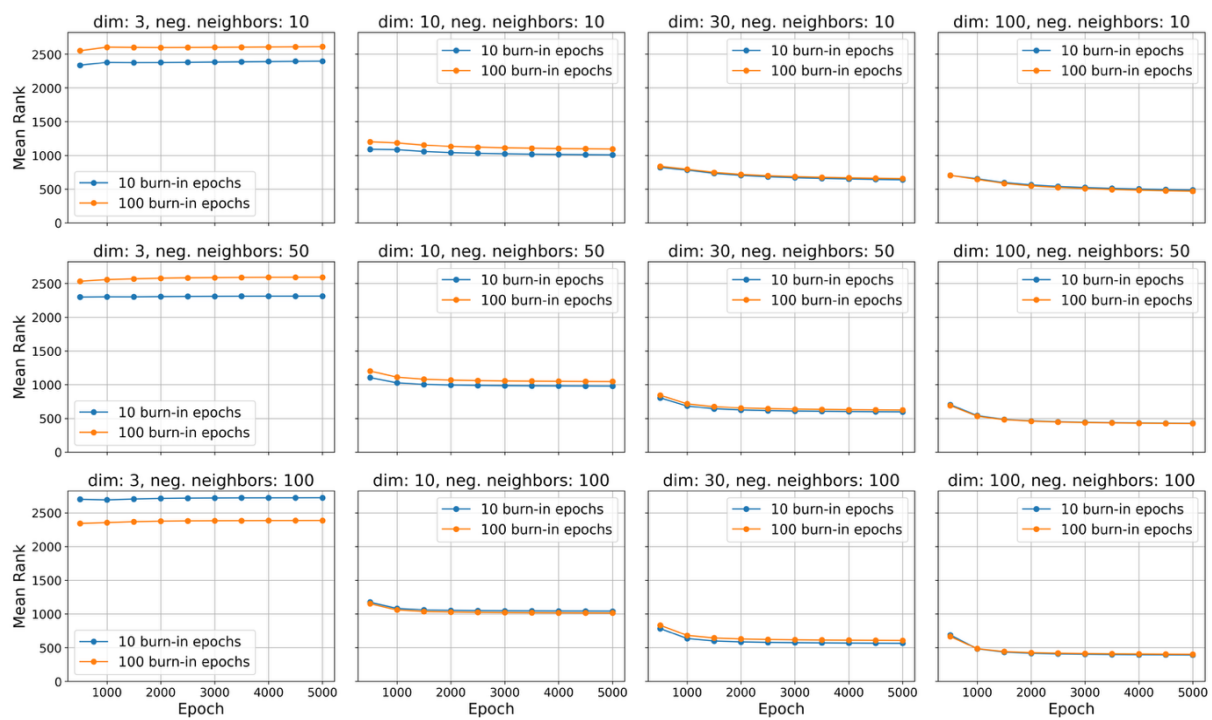

Figure A. Poincaré embeddings trained on undirected SNOMED graph for the hyperparameter space explored.